\documentclass{article}

\usepackage{arxiv}

\usepackage[utf8]{inputenc} 
\usepackage[T1]{fontenc}    
\usepackage{hyperref}       
\usepackage{url}            
\usepackage{booktabs}       
\usepackage{amsfonts}       
\usepackage{nicefrac}       
\usepackage{microtype}      
\usepackage{cleveref}       
\usepackage{lipsum}         
\usepackage{graphicx}
\usepackage{natbib}
\usepackage{doi}

\title{Whisper Turns Stronger: Augmenting Wav2Vec 2.0 for Superior ASR in Low-Resource Languages}

\usepackage{authblk}

\newbox{\orcid}\sbox{\orcid}{\includegraphics[scale=0.06]{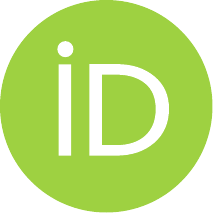}} 
\author[3,1]{%
	\href{https://orcid.org/0000-0002-7673-6511}{\usebox{\orcid}\hspace{1mm}Or Haim Anidjar\thanks{\texttt{orchaimanidjar@gmail.com }}}%
}
\author[4]{%
	\href{https://orcid.org/0000-0002-3815-2582}{\usebox{\orcid}\hspace{1mm}Revital  Marbel\thanks{\texttt{revi85@gmail.com}}}%
}
\author[1,2]{%
	\href{https://orcid.org/0000-0001-5180-4541}{\usebox{\orcid}\hspace{1mm}Roi  Yozevitch\thanks{\texttt{yozevitch@gmail.com}}}%
}
\affil[1]{Department of Computer and Software Engineering, Ariel University, Ariel, Israel}
\affil[2]{Department of Electrical Engineering, Ariel University, Ariel, Israel}
\affil[3]{School of Computer Science, College of Management, Rishon Le'Tzion, Israel}
\affil[4]{Department of Computer Science, Holon Institue of Technology, Holon, Israel}

\hypersetup{
pdftitle={Whisper Turns Stronger: Augmenting Wav2Vec 2.0 for Superior ASR in Low-Resource Languages },
pdfsubject={ASR},
pdfauthor={Or Haim Anidjar, Revital Marbel, Roi Yozevitch},
}

\begin{document}
\maketitle

\begin{abstract}
	Approaching Speech-to-Text and Automatic Speech Recognition problems in low-resource languages is notoriously challenging due to the scarcity of validated datasets and the diversity of dialects. Arabic, Russian, and Portuguese exemplify these difficulties, being low-resource languages due to the many dialects of these languages across different continents worldwide. Moreover, the variety of accents and pronunciations of such languages complicate ASR models' success. With the increasing popularity of Deep Learning and Transformers, acoustic models like the renowned Wav2Vec2 have achieved superior performance in the Speech Recognition field compared to state-of-the-art approaches. However, despite Wav2Vec2's improved efficiency over traditional methods, its performance significantly declines for under-represented languages, even though it requires significantly less labeled data. This paper introduces an end-to-end framework that enhances ASR systems fine-tuned on Wav2Vec2 through data augmentation techniques. To validate our framework's effectiveness, we conducted a detailed experimental evaluation using three datasets from Mozilla's Common Voice project in Arabic, Russian, and Portuguese. Additionally, the framework presented in this paper demonstrates robustness to different diacritics. Ultimately, our approach outperforms two previous baseline models, which are the pre-trained Wav2Vec2 and the well-known Whisper ASR model, resulting in an average relative improvement of 33.9\% in Word Error Rate and a 53.2\% relative improvement in Character Error Rate.
\end{abstract}


\section{Introduction}\label{introduction}

Automatic Speech Recognition (ASR)~\cite{avci2006speech, zoughi2020adaptive, li2022recent} is a technique that converts human speech into readable text. ASR systems, also known as Speech-to-Text (S2T) or transcription systems~\cite{ronao2016human}, are prevalent in applications like virtual assistants such as Apple’s Siri~\cite{mccrocklin2022salukispeech} and Amazon’s Alexa~\cite{brauer2022design}, which heavily rely on ASR systems. The field of speech recognition, especially ASR, has seen exponential growth over the past two decades, becoming popular in various industries such as call centers~\cite{ha2020clovacall}, education~\cite{mccrocklin2022salukispeech, brauer2022design}, finance~\cite{bandi2022artificial}, and healthcare~\cite{sezgin2022editorial}.
The widespread adoption of ASR technology is driven by its ability to enhance user experiences, improve accessibility, and increase operational efficiency across various domains. In call centers, ASR systems streamline customer service operations by providing automated responses and assisting human agents with real-time transcription and sentiment analysis~\cite{ha2020clovacall}. In education, ASR facilitates learning by enabling automatic captioning of lectures, supporting students with disabilities, and assisting language learners~\cite{mccrocklin2022salukispeech, brauer2022design}. In finance, ASR improves productivity by automating transcription of financial meetings and ensuring compliance through accurate documentation of verbal communications~\cite{bandi2022artificial}. In healthcare, ASR aids in clinical documentation, allowing healthcare professionals to focus more on patient care by reducing the time spent on paperwork~\cite{sezgin2022editorial}.

Given the increasing demand for this technology, this paper proposes an ASR framework for under-represented languages, which have limited resources and data for developing accurate ASR systems. Such a framework needs to be robust to a variety of dialects~\cite{dorn2019dialect}, languages, and real-world sound quality, particularly in noisy environments~\cite{kinoshita2020improving}. The scarcity of validated datasets and the diversity of dialects in under-represented languages like Arabic, Russian, and Portuguese present significant challenges for ASR development. These languages exhibit a wide range of accents and pronunciations, further complicating the task of creating effective ASR models.

Recent advancements in speech recognition have significantly improved transcription accuracy~\cite{baevski2020wav2vec}, a.k.a Wav2Vec2 model. Techniques involving deep learning and self-supervised learning (SSL) have revolutionized the field by allowing models to learn from large amounts of unlabeled data, thereby reducing the dependency on labeled datasets. Notably, the Wav2Vec2 model~\cite{baevski2020wav2vec} has achieved superior performance by leveraging SSL to pre-train on vast amounts of audio data before fine-tuning on specific tasks. However, challenges such as background noise and varying accents within the same language still pose difficulties. For instance, in Arabic, accents vary greatly depending on the speaker's country of origin, impacting the performance of ASR systems that may have been trained on a limited set of accents.

It has been demonstrated that pre-trained models, fine-tuned on minimal labeled speech data, achieve competitive results with state-of-the-art ASR systems~\cite{baevski2020wav2vec}. For example, the Wav2Vec2 model~\cite{baevski2020wav2vec}, using only a few hours of labeled data, achieves a Word Error Rate (WER)~\cite{deleglise2009improvements} of less than 5\% on the clean test set of the LibriSpeech dataset~\cite{garnerin2021investigating}. Despite its effectiveness, Wav2Vec2 may underperform in under-represented languages, as evidenced by the challenges with the Arabic language.

\subsection{Arabic Dialects}

Arabic, a Semitic~\cite{huehnergard2019introduction} language spoken by over 300 million people across the Middle East, North Africa, and Asia, has various dialects. While Modern Standard Arabic (MSA)~\cite{qwaider2019can} is used for formal communication, spoken Arabic varies significantly across regions~\cite{alhelbawy2020nlp}. These dialects, which some consider different languages, have distinct features and characteristics. This variation is not unique to Arabic and is also present in European dialects~\cite{khosravani2021modeling, alsayadi2022dialectal}.

Pronunciation~\cite{caballero2014evolutionary} of certain consonants and vowels varies widely among dialects. For example, the \textit{'d'} sound in Standard Arabic may be pronounced as \textit{'z'} in some dialects, and the \textit{'q'} sound as \textit{'g'} in others. These differences can hinder mutual understanding among speakers of different dialects.

Vocabulary differences also pose a challenge. Although many words are shared, some are specific to particular dialects. This affects S2T system performance, as it may not recognize dialect-specific words. For example, an ASR system trained on a dialect using "\textit{Automobil}" for "car" may not accurately transcribe "\textit{sayyara}" from another dialect. Robustness in S2T systems requires accounting for these dialectal differences.

Given these challenges, training a state-of-the-art ASR model like Wav2Vec2 on diverse dialects and vocabulary is crucial. This can be achieved with large, diverse datasets and advanced machine learning algorithms. Such training enables the ASR system to recognize a variety of words and phrases accurately across different Arabic dialects. To evaluate our framework's performance, we use both WER~\cite{deleglise2009improvements} and Character Error Rate (CER)~\cite{hou2020large, kumar2022comprehensive}, a similar metric that measures errors at the character level.

\textbf{Comparison with Existing Methods.} Our study diverges from the approaches taken in SpecWav2vec-F~\cite{10587399} and the Wav2Vec-Aug~\cite{sriram2022wav2vec} model in several notable ways. While SpecWav2vec-F enhances low-resource speech recognition through a unique SpecAugment-based technique applied to datasets such as KSC for English, Chinese, and Kazakh, our work specifically targets languages with high dialectal variation, such as Arabic, Russian, and Portuguese. These languages, represented in the Mozilla Common Voice dataset, present distinctive challenges due to their spontaneous, real-world nature, unlike the KSC corpus and LibriSpeech datasets. LibriSpeech, created from controlled audiobook recordings, lacks the variability in pronunciation and accent found in natural speech, which is central to our investigation. Consequently, our study provides insight into ASR model performance under more realistic conditions than previous work, addressing dialectal variability as a central challenge.

Moreover, a key distinction in our methodology is the inclusion of both WER and CER as evaluation metrics, recognizing CER’s critical importance for dialect-heavy languages such as Arabic. Given the influence of regional dialects on ASR accuracy, CER provides a more granular analysis of errors that reflects the nuanced phonetic and lexical differences among dialects. While SpecWav2vec-F~\cite{10587399} and the Wav2Vec-Aug~\cite{sriram2022wav2vec} models report WER results, our framework demonstrates a 33.9\% improvement in WER and a 53.2\% improvement in CER, outperforming both standard Wav2Vec2 and Whisper ASR models. This comprehensive error analysis highlights our model's robustness in managing dialectal variation across low-resource languages, making our approach particularly well-suited for ASR systems facing real-life, diverse linguistic landscapes.

\subsection{Our Contribution}

This paper presents a robust method for fine-tuning the Wav2Vec2 speech recognition model using a novel data augmentation approach. The algorithm was trained and tested on three under-represented languages: Arabic, Portuguese, and Russian. These languages were chosen due to their low data availability. The fine-tuning process used limited audio data (Arabic and Portuguese: 17 hours, Russian: 30 hours), resulting in an average WER improvement of 34\% and a CER improvement of over 50\%. The results demonstrate robustness to different languages with varied grammatical rules and syntax.

First, our approach addresses the challenge of data scarcity in under-represented languages by employing advanced data augmentation techniques. This enables the creation of more diverse training datasets, significantly enhancing the model's performance despite the limited available data.

Second, we validated our framework's effectiveness through extensive experimental evaluation, demonstrating substantial improvements in both WER and CER metrics. This empirical evidence underscores the potential of our method to improve ASR accuracy across different language families, highlighting its generalizability.

Third, we compared our augmented Wav2Vec2 model against two prominent baseline models: the pre-trained Wav2Vec2 and the well-known Whisper ASR model. Our approach outperformed these baselines, resulting in an average relative improvement of 33.9\% in WER and a 53.2\% relative improvement in CER, thus establishing a new benchmark for ASR performance in low-resource languages.

Finally, our framework demonstrates robustness to the variations in dialects and pronunciations within each tested language. By accounting for these linguistic nuances, our method enhances the adaptability of ASR systems to real-world scenarios where speakers exhibit diverse accents and dialects. This contribution is crucial for developing more inclusive and effective speech recognition technologies.

Table \ref{tab:list_of_abbreviations} lists the abbreviations used throughout this paper.

\begin{table}[!ht]
	\centering
    \begin{tabular}  {|p{1.7cm}|p{4.3cm}|} \hline
    	\bf {Abbreviation} & \bf {Meaning} \\\hline
    	ASR & Automatic Speech Recognition \\
            \hline
            CER &  Character Error Rate\\
            \hline
            CTC & Connectionist Temporal Classification \\
            \hline
            DER & Diarization Error Rate \\
            \hline
            S2T & Speech-2-Text \\
            \hline
            SCD & Speaker Change Detection \\
            \hline
            SD & Speaker Diarization \\
            \hline
            SR & Speech Recognition \\
            \hline
            SSL & Self Supervised Learning \\
            \hline
            WER & Word Error Rate \\
            \hline
    \end{tabular}\par
    	\caption{table}{List of Abbreviations.} \label{tab:list_of_abbreviations}
\end{table}

\section{Related Work} \label{sec:related_work}

Representation Learning~\cite{bengio2013representation} involves techniques for vector representation of data, used for tasks like classification and clustering~\cite{7528397}. It can replace manual feature extraction and engineering. Transformers, a type of representation learning, use Self Supervised Learning (SSL)~\cite{nguyen2021structural, jaiswal2020survey, shurrab2022self} and self-attention~\cite{zhai2019s4l} to learn optimal representations of raw data for specific tasks. SSL trains models with unlabeled data before fine-tuning with labeled data, addressing the scarcity of labeled data~\cite{babbar2019data}. This approach enables learning general data representations from unlabeled examples, later refined with labeled data by adding a predictor to the model.

Wav2Vec2~\cite{baevski2020wav2vec, NEURIPS2020_92d1e1eb} is a Transformer-based~\cite{lin2022survey} model used in speech recognition, leveraging SSL. It processes raw audio waveforms to generate vector-based language representations~\cite{sasajima1996representation}. Wav2Vec2 is now considered state-of-the-art for its high-accuracy speech transcription and ability to handle various languages, accents, and speech styles.

Previous research on Wav2Vec2 has focused on enhancing its performance through techniques like fine-tuning, modifying the model architecture or hyperparameters, and incorporating additional training objectives. For example, the Wav2Vec2-xlsr-53~\cite{deschamps2022investigating} model, trained on 53 languages, achieved state-of-the-art performance on multiple speech recognition tasks~\cite{shahgir2022applying}. Other studies~\cite{farias2022bilingual} have explored Wav2Vec2 for tasks such as speaker identification~\cite{malek2022target}, language identification~\cite{chakravarthi2022dravidiancodemix}, and keyword spotting~\cite{ahmed2022towards}. Wav2Vec2's widespread adoption in the speech recognition community is largely due to the advantages of SSL.

Training and fine-tuning speech recognition models with limited data~\cite{thomas2020transliteration} often result in models with higher error rates. ASR models typically require large, diverse datasets for efficient learning. This gap in robustness is the main focus of this paper, which addresses ASR systems for low-resource and under-represented languages~\cite{shor2019personalizing}. Limited data can lead to overfitting, where models perform well on training data but poorly on new data. Data augmentation~\cite{shahnawazuddin2020creating} is one suggested method to mitigate this issue~\cite{alsayadi2021data}.

Another approach involves clustering unlabeled data~\cite{bakheet2021improving, hsu2021hubert}, which is easier to obtain. However, these methods do not guarantee low error rates, raising concerns about ASR system reliability when WER~\cite{deleglise2009improvements} is high.

Voice models are sensitive due to high variance and noisy data. ASR models face additional complexity from language processing layers. High WER values do not necessarily indicate failure. For example, \cite{anidjar2023speech} developed an end-to-end framework for Speaker Change Detection (SCD)~\cite{anidjar2020thousand, meng2017hierarchical, hruz2017convolutional} and Speaker Diarization (SD)~\cite{lin2019lstm, shum2013unsupervised, silnova2020probabilistic}, achieving 97.66\% F1-Score for SCD and 10.28 DER~\cite{deleglise2009improvements} despite a 40.3\% WER in English ASR.

\cite{kepuska2017comparing} designed a tool to compare commercial ASR systems like Microsoft Speech API~\cite{kepuska2017comparing} and Google Speech API~\cite{anggraini2018speech} with open-source systems like Sphinx-4~\cite{walker2004sphinx}. Despite Sphinx-4's 37\% WER~\cite{li2020towards, nakatani2019improving}, it remains competitive for speech recognition tasks~\cite{walker2004sphinx, hafeez2014speaker}, compared to Microsoft Speech API's 18\% WER and Google Speech API's 9\% WER.

\cite{radford2022robust} explored ASR systems trained to predict transcriptions from large-scale audio recordings. Their model, scaled to 680,000 hours of multilingual and multitask supervision, achieved 9.9\% WER in English and 29.2\% WER in a multilingual dataset, using weak supervision~\cite{kuang2022firebolt} on the Wav2Vec2~\cite{baevski2020wav2vec, NEURIPS2020_92d1e1eb} architecture.

\cite{tran2022towards} presented a speech-representation anonymization framework using selective noise perturbation for privacy and security in cloud-based ASR or Speech Emotion Recognition (SER)~\cite{guo2022learning}. Even with a 39.6\% WER, their framework effectively recognized emotions in audio recordings.

\subsection{Whisper}

Recently, \href{https://open.ai/}{Open.AI} launched Whisper~\cite{radford2022robust, radford2023robust}, a general-purpose speech recognition model trained on diverse audio data. It is a multitask model capable of multilingual speech recognition, translation, and language identification~\cite{almeida2014feature}. On the Fleurs dataset~\cite{conneau2022fleurs}, Whisper achieved a 16\% WER in Arabic. Whisper, with over 1.5 billion parameters, is significantly larger than Wav2Vec2, which has around 3 million parameters. However, two key differences must be considered when comparing Whisper and Wav2Vec2:

\textbf{(1).} Whisper's WER is calculated using a custom-built text normalizer before WER computation, improving accuracy by correcting common mistakes and normalizing transcribed text to match the reference text.

\textbf{(2).} This method, while effective, may not reflect real-world scenarios where training data is often flawed. For example, training data might include silent audio recordings labeled with full sentences.

\section{Datasets} \label{sec:datasets}

The Mozilla Common Voice dataset \cite{ardila2019common, berkson2019building, chachadi2022voice} is a free collection of recorded speech data. It is publicly available and fosters innovation, making it a key resource for commercial machine-learning competitions based on speech technology. Common Voice is a multilingual dataset and is one of the largest publicly available voice datasets of its kind.

Common Voice is used to both train and evaluate speech recognition algorithms. The dataset contains recordings from over 400,000 people speaking multiple languages, including a variety of voices across different ages, genders, accents, and speaking styles. It includes diverse transcriptions such as informal conversations, news articles, and public service announcements. However, not all languages are equally represented. For instance, the English language has over 2,000 hours of audio data, whereas Arabic has only 89 hours. This work focuses on ASR systems for low-resource languages, specifically the following three underrepresented languages: Arabic, Russian, and Portuguese.

\begin{itemize}
    \item \textbf{Arabic.} The Arabic version of the Common Voice dataset consists of approximately 89 hours of community-validated audio, 147 hours of total audio, and 1,309 unique voices in mp3 format. Notably, \textbf{the train split only contains 17 hours of labeled data.}

    \item \textbf{Portuguese.} The Portuguese part of the Common Voice dataset includes 126 hours of community-validated audio, 151 hours of total audio, and 2,621 unique voices in mp3 format. \textbf{The train split only contains 17 hours of labeled data.}

    \item \textbf{Russian.} The Russian part of the Common Voice dataset includes 180 hours of community-validated audio, 215 hours of total audio, and 2,731 unique voices in mp3 format. \textbf{The train split only contains 30 hours of labeled data.}
\end{itemize}

Unfortunately, the Common Voice dataset contains some samples with empty audio recordings (silent parts) that still have labels, or labels that do not match the audio, leading to misclassifications and affecting the WER and CER.

\subsection{Data Pre-Processing}

To prepare the data for training, the Wav2Vec2 framework requires the following steps for the audio data:

\begin{itemize}
    \item Change the sample rate from 44kHz to 16kHz, as Wav2Vec2 works with this sample rate.
    \item Remove special characters.
    \item Remove punctuation marks.
\end{itemize}

\section{Augmentation Methods} \label{sec:aug_methods}
Data augmentation~\cite{shahnawazuddin2020creating} is a technique used to artificially increase the size of a dataset by creating modified versions of existing data. For audio, this can involve transformations such as pitch shifting, time stretching, or adding noise. Data augmentation improves model performance and generalization by providing additional training examples. The augmented data remains viable as long as the augmentations do not interfere with the transcription of the original audio~\cite{ragni2014data}. The new data can be used to further train the model.

Focusing on a minimal number of high-impact augmentations is crucial for improving the model effectively. Insufficient augmentation can cause overfitting due to a lack of data variety, while excessive augmentation can lead to underfitting, losing critical data and making words unrecognizable. Increasing data variance and robustness is essential for significant improvements. In this work, three augmentation methods are considered:

\begin{itemize}
    \item Band-Stop~\cite{thai2019synthetic} (Section~\ref{sec:band_stop}).
    \item Gaussian-Noise~\cite{scharenborg2017building} (Section~\ref{sec:gaussian_noise}).
    \item Pitch-Shift~\cite{scharenborg2017building, thai2019synthetic} (Section~\ref{sec:pitch_shift}).
\end{itemize}

Finally, Section~\ref{sec:aug_combination} discusses the combination of augmentation methods in relation to model construction.

\subsection{Band Stop} \label{sec:band_stop}
\textbf{Band-Stop}~\cite{roonizi2021band} augmentation involves removing a specific frequency range from an audio signal using a band-stop filter, which attenuates frequencies within a certain range and allows others to pass through. Human hearing ranges between 0-4000Hz, so these frequencies are used as the minimum and maximum thresholds. The min/max cutoff range is represented by the bandwidth fraction (the absolute bandwidth divided by the center frequency, represented between 0-200\%), indicating the relative portion of the frequency spectrum to cut off. The steepness of the cutoff is set in dB. This augmentation simulates various (Arabic) accents.

\subsection{Gaussian-Noise} \label{sec:gaussian_noise}
Adding \textbf{Gaussian-Noise}~\cite{el2020blind} to training data can improve an ASR model's robustness and generalization. This method helps the model handle variations in input data, such as different accents or speaking styles, and real-world scenarios with noisy data. To augment audio recordings, an array of the same shape as the audio is created with random samples from a uniform distribution over 0.001 to 0.03 Hz. The amplitude of the original audio is multiplied by this array, and the result is added to the original amplitude to create the augmented file.

\subsection{Pitch Shift} \label{sec:pitch_shift}
\textbf{Pitch Shift}~\cite{gfeller2020spice} is achieved by uniformly shifting the tempo of the entire audio recording by a certain number of semitones, synthesizing different sounding voices. Many speech datasets have a small variety of speakers, with each recording contributing significantly. The Arabic Common Voice dataset has roughly 15 different speakers per hour. Adding variance can significantly improve model generalization and prevent overfitting. For each recording, the semitones are randomly chosen from the range $[-6, 6]$.

\subsection{Augmentations Combination} \label{sec:aug_combination}
These three augmentations differ inherently. Pitch-Shift uniformly changes the semitone of the entire recording, Band-Stop removes a specific frequency range, and Gaussian-Noise adjusts the amplitude to create white noise. Each plays a crucial role in improving the model's accuracy.

\section{Framework} \label{sec:our_approach}

\subsection{Model Architecture}

This paper examines the effectiveness of fine-tuning the Wav2Vec2-xlsr-53 model~\cite{deschamps2022investigating} on augmented data. The Wav2Vec2-xlsr-53 model, pre-trained on 53 languages by the team at \href{https://ai.facebook.com/}{Facebook AI Research} in September 2020, represents a state-of-the-art approach for converting raw audio waveforms into high-quality text representations. It is based on the concept of SSL, where the model learns to predict missing segments of the input waveform.

The model's architecture consists of three main components:

\begin{itemize}
    \item \textbf{Pre-Processing.} The model receives a raw audio matrix as input and outputs latent speech representations for each time step among T time steps.
    
    \item \textbf{Speech-Encoding.} The speech representations are fed into a Transformer that creates T representations, extracting information from the sequence.
    
    \item \textbf{CTC-Clustering.} The feature encoder output is discretized to represent the targets (outputs) using a self-supervised objective function.
\end{itemize}

\begin{figure*}
    \centering
    \includegraphics[scale=0.25]{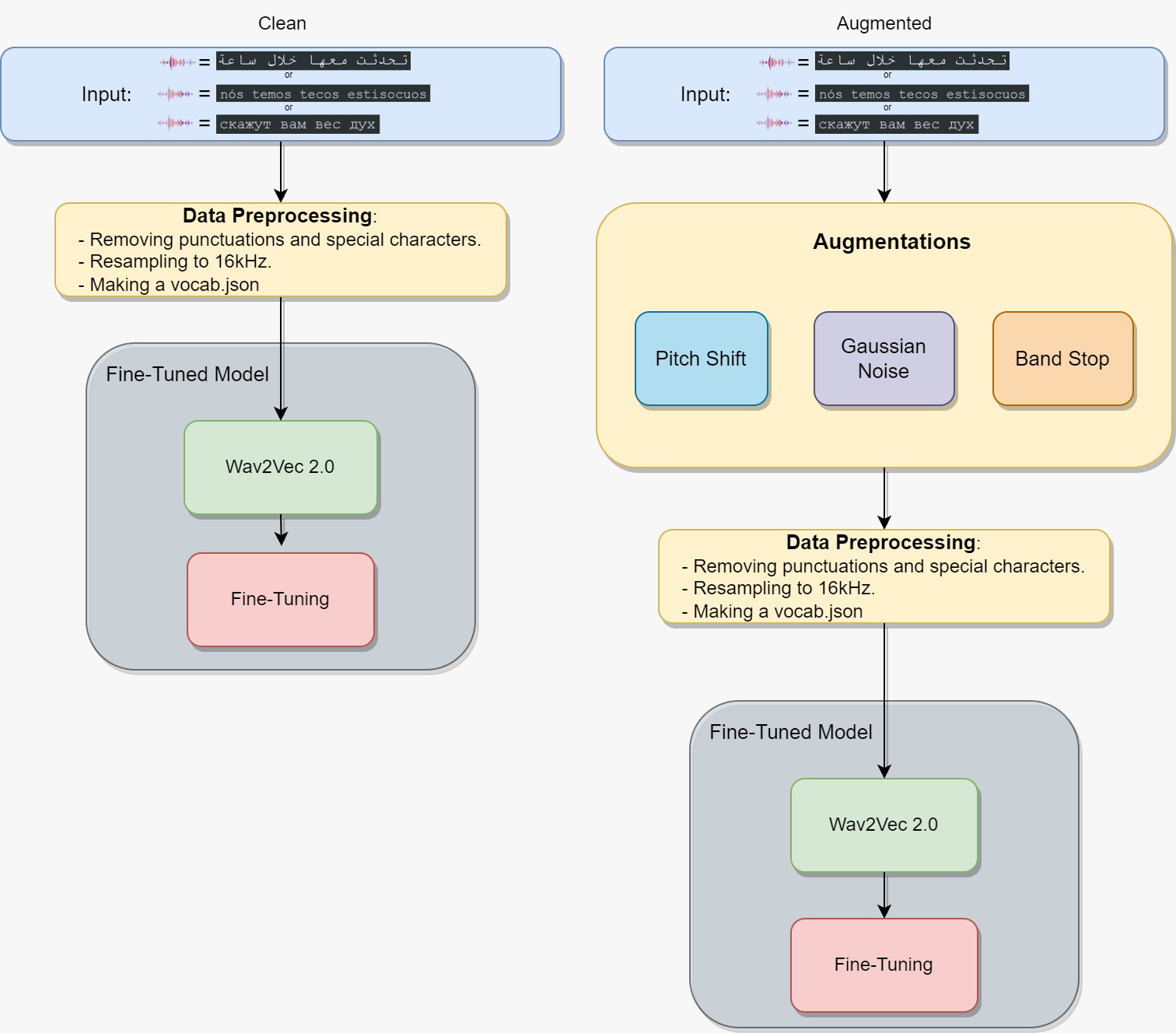}
    \caption{The framework presented in this paper.}
    \label{fig:model_flow}
\end{figure*}

The Wav2Vec2 model is composed of a multi-layer convolutional feature encoder. The feature encoder includes a temporal convolution followed by a normalization layer and a GELU~\cite{hendrycks2016gaussian} activation function. The encoder's total stride determines the number of T time steps, which serve as the Transformer's input. The Transformer then produces contextualized speech representations. The feature encoder output is fed into a context network that follows the Transformer architecture as described in~\cite{devlin2018bert}. Finally, Wav2Vec2 is trained using the CTC~\cite{higuchi2022hierarchical} loss, as the S2T problem involves sequence alignment. Unlike fixed positional embeddings~\cite{devlin2018bert}, which encode absolute positional information, Wav2Vec2 uses a convolutional layer that acts as a relative positional embedding.

{\bf CTC loss-function.} In ASR systems, aligning each character to its proper location in an audio recording is challenging. The CTC loss computes the loss between a continuous and unsegmented acoustic time-series signal and a target sequence-based label represented by characters. This computation sums over the probability distribution of possible alignments between the speech signal and the textual sequence label, producing a differentiable loss value with respect to each input node. The alignment is assumed to be “many-to-one,” requiring the textual sequence label length to match the input length. We hypothesize that fine-tuning the model on augmented data will improve its performance compared to fine-tuning on clean data.

We evaluated the performance of the fine-tuned models using common ASR metrics, WER and CER (Section~\ref{sec:precision_metrics}), assessing their ability to accurately transcribe speech, capture the content of the input audio, and generalize to new, unseen data. The results showed that fine-tuning the Wav2Vec2-xlsr-53 model on augmented data improved its performance compared to fine-tuning on clean data.

Finally, the model architecture and flow are presented in Figure~\ref{fig:model_flow}.

\subsection{ASR Precision Metrics} \label{sec:precision_metrics}

WER and CER are the most common metrics for evaluating ASR system performance. They compare the system's output to a reference transcription of the same input. The main difference between WER and CER is the unit of measurement: WER is based on the number of incorrect \textit{words} in the system's transcription, while CER is based on the number of incorrect \textit{characters}. The CER formula is given by Eq.(\ref{eq:cer}):

\begin{equation} \label{eq:cer}
    CER = \frac{(I + S + D)}{N} \times 100
\end{equation}
    
where:

\begin{itemize}
    \item $I$ is the number of insertions (characters in the system's transcription but not in the reference transcription).
    
    \item $S$ is the number of substitutions (characters in the system's transcription that differ from those in the reference transcription).
    
    \item $D$ is the number of deletions (characters in the reference transcription but not in the system's transcription).
    
    \item $N$ is the total number of characters in the reference transcription.
\end{itemize}

The WER formula is similar to CER and is given by Eq.(\ref{eq:wer}):

\begin{equation} \label{eq:wer}
    WER = \frac{(I + S + D)}{N} \times 100
\end{equation}

where:

\begin{itemize}
    \item $I$ is the number of incorrect words in the system's output.
    
    \item $S$ is the number of words correctly recognized but out of order in the system's output.
    
    \item $D$ is the number of words deleted from the reference transcription.
    
    \item $N$ is the total number of words in the reference transcription.
\end{itemize}

\subsection{Arabic Diacritics} \label{Arabic_diac}

\begin{figure*}
    \centering
    \includegraphics[scale=0.57]{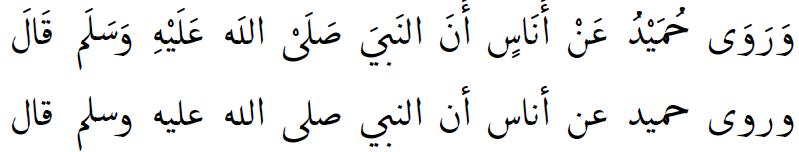}
    \caption{An example of two identical sentences using different diacritics.}
    \label{fig:arab_diacritics_fig}
\end{figure*}

Arabic diacritics are 11 symbols added to letters in the Arabic alphabet to indicate vowel sounds and other phonetic features in specific words. For example, consider the two identical sentences in Figure~\ref{fig:arab_diacritics_fig}. These diacritics significantly affect the reported WER; even a single error in one diacritic can cause the entire word to be incorrect. Thus, diacritics-based WER computation does not accurately reflect the model's performance.

\subsubsection{WER vs CER} \label{wer_vs_cer}

Generally, WER is more commonly used than CER because it is easier to understand and interpret. However, CER can be useful in certain situations, such as when the transcription includes proper nouns or difficult-to-spell words or when it contains many homophones (words that sound the same but are spelled differently). Moreover, if the system systematically fails to recognize spaces between words, CER and WER metrics will produce different results. For example, consider the following two Portuguese sentences:

\begin{quote}
    Reference: \textbf{é necessário fornecer quando formulado uma avaliação}
    
    Prediction: \textbf{e necessário ponecer quando forme lado u mavalação}
\end{quote}

The WER between the reference and prediction sentences is $85.7\%$, while the CER between them is only $17.3\%$.

\section{Experimental Evaluation and Results} \label{sec:results}

This section presents the WER and CER results for the proposed model across the three tested languages. A considerable improvement was achieved across all languages. In all experiments, the following training parameters were used: 500 warm-up steps, a learning rate of $3e-4$, a batch size of 16, and evaluation every 100 steps. The full model details, as well as the training process (layers size and type, learning rate, dropout layers, loss function, and model size), are as provided in Section 3 in~\cite{baevski2020wav2vec}.

\subsection{Arabic Results}

Throughout the experiment, the Wav2Vec2-xlsr-53 model was fine-tuned on a combination of clean and augmented data. As explained in Section \ref{sec:aug_methods}, the chosen augmentations were pitch shift, Gaussian noise, and band-stop filter. To test our hypothesis that data augmentation is highly beneficial for ASR systems, we first trained the Wav2Vec2-xlsr-53 model on 17 hours of clean audio data (Common Voice 11.0) for the Arabic language only, since it is the most complex and dialect-rich. To determine the best augmentation-based combination (Section~\ref{sec:aug_combination}), the model was fine-tuned (train-split only) in the following manners:

\textbf{(1).} 100\% clean data + 20\% augmented data, with the model trained using 20\% more augmented data, applied three times for the three different augmentations (Sections~\ref{sec:band_stop},~\ref{sec:gaussian_noise}, and~\ref{sec:pitch_shift}). For each model, the additional 20\% augmented data was chosen randomly from the train split. This step trains and evaluates three different models.

\textbf{(2).} 100\% clean data + 20\% augmented data from every pair of augmentations. This includes (i) one model with 20\% band-stop and 20\% Gaussian-noise augmentations; (ii) one model with 20\% band-stop and 20\% pitch-shift augmentations; and (iii) one model with 20\% Gaussian-noise and 20\% pitch-shift augmentations. For each model, the 40\% augmented data (20\% of one augmentation method and 20\% of another) was chosen randomly from the train split. This step trains and evaluates three different models.

\textbf{(3).} 100\% clean data + 20\% from each augmentation. This model includes 20\% band-stop augmentation, 20\% Gaussian-noise augmentation, and 20\% pitch-shift augmentation. For this model, the 60\% augmented data (20\% of each augmentation) was chosen randomly from the train split. This step trains and evaluates one model.

\begin{figure*}
    \centering
    \includegraphics[scale=0.15]{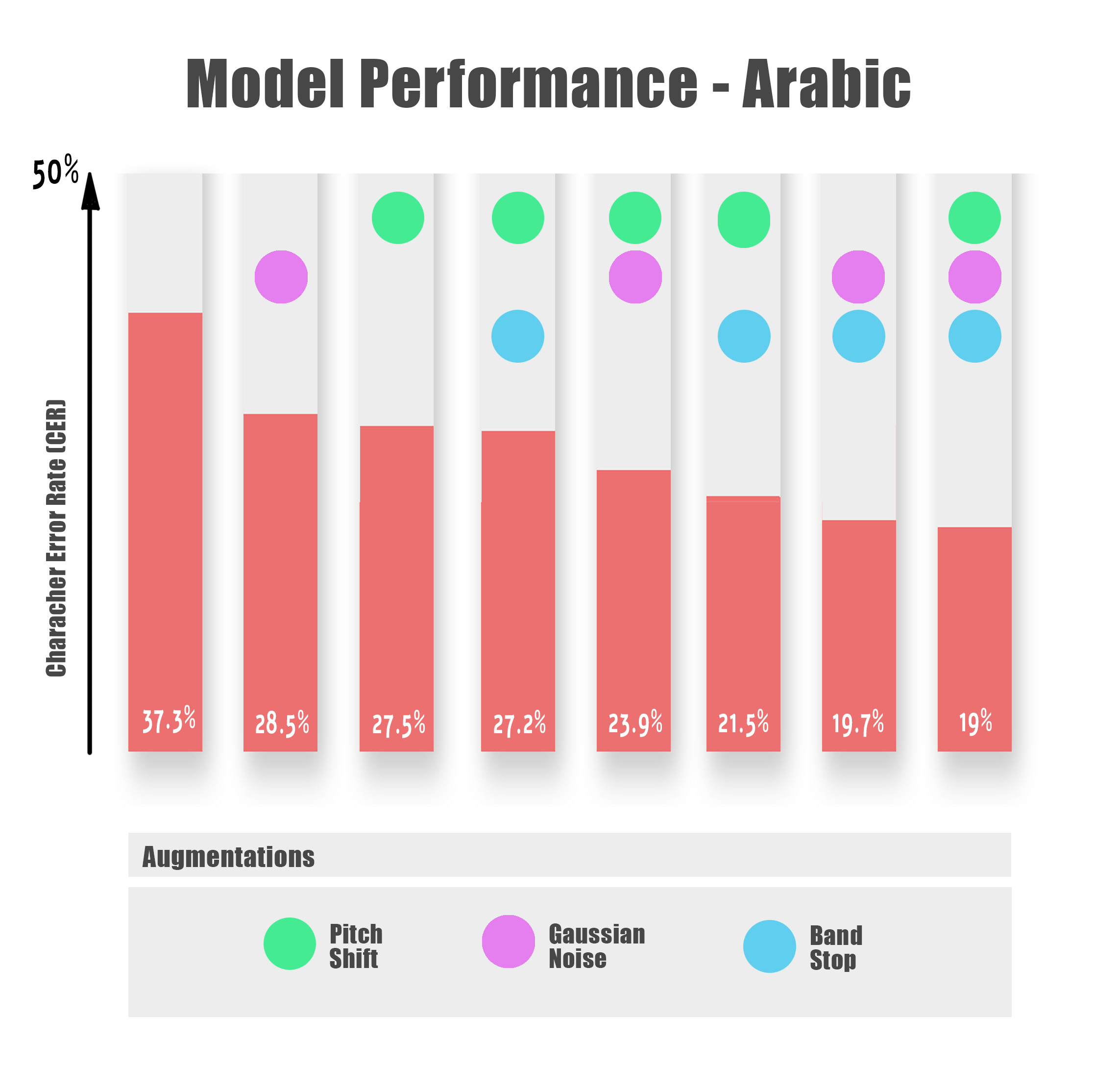}
    \caption{CER bar-chart for different combinations of augmentation methods. It is clear that the more augmentations used, the better the result. The best combination is the use of all three augmentations: Pitch-Shift, Gaussian-Noise, and Band-Stop, achieving a 19\% CER for a variety of diacritics in Arabic, which triples the size of the letters vocabulary to 80 letters.}
    \label{fig:arabic_results}
\end{figure*}

The CER bar-chart for all seven models is illustrated in Figure~\ref{fig:arabic_results}. As shown, using augmented data significantly improves the model's performance, with the best result (CER = 19.0\%) achieved using all three augmentations. The more augmentations utilized, the better the model performs. The CER of the non-augmented data is 37.5\%. Additionally, it is evident that a single augmentation is inferior to any combination of two augmentations, and using all three is superior to all others.

\subsection{Comparing Arabic, Russian and Portuguese}

Based on the results for the Arabic language, two additional languages, Russian and Portuguese, were tested. While Arabic is a Semitic language, Portuguese and Russian are Latin and Slavic, respectively. Testing on these languages will demonstrate the model's generalization across different language families. Additionally, these languages have relatively small training datasets, as presented in Section~\ref{sec:datasets}. The Portuguese training data consists of approximately 17 hours of audio recordings, and the Russian training data consists of approximately 30 hours of audio recordings (train-splits from the Common Voice dataset). The WER and CER results are summarized in Tables~\ref{table_wer} and~\ref{table_cer}:

\begin{table*}[h]
    \centering
    \begin{tabular}{ |p{2cm}||p{2cm}|p{2cm}|p{2cm}| }
        \hline
        \multicolumn{4}{|c|}{Word Error Rate (WER)} 
        \\
        \hline
         Language& Clean data [\%]& Augmented [\%]&Improvement [\%]\\
         \hline
        
        Arabic & 46.5		& \textbf{27.6}		& 40.65	\\
        Russian & 54.6		& \textbf{35.8}		& 34.43	\\
        Portuguese & 43.3   & \textbf{31.8}		& 26.56	\\
        \hline 
        \end{tabular}
        \vspace{10pt}
        \caption{Results table of WER for all languages. The highlighted results in the \textit{Augmented} column demonstrate significant improvements over the \textit{Clean Data} column. Specifically, the model's performance on Arabic data improved relatively by 40.65\%, i.e. reducing the WER from 46.5\% to 27.6\%. Similarly, the Russian data showed a 34.43\% relative improvement, with the WER decreasing from 54.6\% to 35.8\%. Lastly, the Portuguese data exhibited a 26.56\% relative enhancement, lowering the WER from 43.3\% to 31.8\%. These improvements highlight the effectiveness of the augmentation techniques across different language families, despite the relatively small training datasets.}
        \label{table_wer}
\end{table*}

Table~\ref{table_wer} shows the final WER for all three tested languages with and without the augmentations proposed in Section \ref{sec:aug_methods}. While the average WER for the clean, non-augmented data is relatively high at approximately 48\%, the average WER for the augmented data is approximately 32\%. This represents an average accuracy improvement of approximately 34\%.

Additionally, Table~\ref{table_cer} shows the CER results for the three languages:

\begin{table*}[h]
    \centering
    \begin{tabular}{ |p{2cm}||p{2cm}|p{2cm}|p{2cm}| }
        
        \hline
        \multicolumn{4}{|c|}{Character Error Rate (CER)} 
        \\
        \hline
         Language& Clean data [\%]& Augmented [\%]&Improvement [\%]\\
         \hline
        
         Arabic   &22.3		& \textbf{9.0}		& 59.64	\\
         Russian &  22.3		& \textbf{10.2}		& 54.26	\\
         Portuguese &21.2	& \textbf{11.5}		& 45.75	\\
        
        \hline
				
        \end{tabular}
        \vspace{10pt}
        \caption{Results table of CER for all languages. The highlighted results in the \textit{Augmented} column demonstrate significant improvements over the \textit{Clean Data} column. Specifically, the model's performance on Arabic data improved relatively by 59.64\%, reducing the CER from 22.3\% to 9.0\%. Similarly, the Russian data showed a 54.26\% relative improvement, with the CER decreasing from 22.3\% to 10.2\%. Lastly, the Portuguese data exhibited a 45.75\% relative enhancement, lowering the CER from 21.2\% to 11.5\%. These improvements underscore the effectiveness of the augmentation techniques across different language families, significantly enhancing character-level accuracy despite the relatively small training datasets.}
     
        \label{table_cer}
\end{table*}

Table~\ref{table_cer} shows the final CER for all three tested languages with and without the augmentations proposed in Section \ref{sec:aug_methods}. The average CER for the clean, non-augmented data is approximately 22\%, while the average CER for the augmented data is approximately 10\%. This represents an average accuracy improvement of approximately 53\%. The reason why CER produces better results was already explained in Section~\ref{wer_vs_cer}.

\subsection{Baseline Models Comparison - Whisper and Wav2Vec2}

This section presents the WER and CER results for the state-of-the-art for multilingual ASR, namely, Whisper model~\cite{radford2023robust} (Whisper-tiny) across the three tested languages. We compare these results to our data-augmented Wav2Vec2 model to highlight the improvements achieved.

\begin{table*}
    \centering
    \begin{tabular}{|c|c|c|}
        \hline
        \textbf{Language} & \textbf{WER (\%)} & \textbf{CER (\%)} \\ \hline
        
        Arabic & 90.9 \textbf{(27.6)} & N/A \\ \hline

        Russian & 40.6 \textbf{(35.8)} & N/A  \\ \hline
        
        Portuguese & 35.2 \textbf{(31.8)} & N/A  \\ \hline
        
    \end{tabular}
    \vspace{10pt}
    \caption{WER and CER results of the Whisper~\cite{radford2023robust} model on Common-Voice dataset, across different languages - Arabic, Russian and Portuguese. The referenced paper (\cite{radford2023robust}) does not report CER value; thus, they were not included in this table. The highlighted results in parentheses are corresponding with the Augmented-Wa2Vec2 model, which represents our approach as in Table~\ref{table_wer}.}
    \label{tab:whisper_results}
\end{table*}

The observed outperformance of our augmented Wav2Vec2 model can be attributed to several key factors. Firstly, the data augmentation techniques employed in our framework significantly enhance the diversity and quantity of training data. This allows the model to better generalize to the various accents and dialects present in low-resource languages such as Arabic, Russian, and Portuguese.

Compared to the baseline models, the Whisper ASR model, our approach leverages advanced augmentation strategies that mitigate the limitations posed by the scarcity of labeled data. The Whisper ASR model, while effective, does not incorporate the same level of fine-tuning and augmentation, which limits its ability to adapt to the nuances of low-resource languages. Similarly, the pre-trained Wav2Vec2 large model, despite its strong baseline performance, struggles with the same issues when applied to under-represented languages without additional augmentation, as presented in the \textit{Clean Data} column in Tables~\ref{table_wer} and~\ref{table_cer}.

Moreover, our model's robustness to different diacritics, which are prevalent in languages like Arabic, further enhances its WER and CER metrics. Diacritics can drastically change the meaning and pronunciation of words, posing significant challenges for ASR models. By incorporating techniques to handle these variations, our framework ensures more reliable recognition across different contexts.

The significant relative improvements in WER and CER underscore the effectiveness of our approach, by two means; 

\textbf{(i)} in Tables~\ref{table_wer} and~\ref{table_cer}, the average relative improvement of 33.9\% in WER indicates that our model makes substantially fewer errors in word recognition compared to the baseline models. Similarly, the 53.2\% average relative improvement in CER highlights the model's enhanced precision in recognizing characters, which is crucial for languages with complex orthographies.

\textbf{(ii)} in Table~\ref{tab:whisper_results}, where our approach's outperformance is outlined for all three languages since the Whisper model's WER for the three of them is significantly higher (the CER metric was not provided in the original Whisper paper~\cite{radford2023robust}).

Eventually, our end-to-end framework for augmenting Wav2Vec2 demonstrates clear advantages over the baseline models, particularly in handling low-resource languages with diverse dialects and pronunciations. The improvements in WER and CER metrics validate the effectiveness of our data augmentation techniques and highlight the potential for further advancements in ASR technology for under-represented languages.

\subsection{Discussion}
The experimental evaluation conducted in this study underscores the significant impact of data augmentation on ASR system performance, compared to the pre-trained Wav2Vec2 model~\cite{baevski2020wav2vec}. Our results clearly demonstrate that augmenting the training data with techniques such as pitch shifting, Gaussian noise, and band-stop filtering leads to substantial improvements in both WER and CER metrics. These enhancements are particularly notable given the limited training data available for underrepresented languages in the Common Voice dataset.

For the Arabic language, which is uniquely challenging due to its complex and dialect-rich nature, our model achieved a remarkable reduction in CER from 37.3\% to 19.0\% using all three augmentation methods. This indicates that augmented data not only enhances the model's ability to generalize across different dialects and pronunciations but also significantly reduces errors related to diacritics.

Applying the same augmentation strategies to Russian and Portuguese, the model continued to show improved performance. The WER for Arabic decreased from 46.5\% to 27.6\%; for Russian, it decreased from 54.6\% to 35.8\% ; and for Portuguese, it decreased from 43.3\% to 31.8\%. These results underscore the robustness of our augmentation techniques across different language families and their potential for application to other low-resource languages.

When compared to the Whisper model~\cite{radford2023robust}, our data-augmented Wav2Vec2 approach demonstrates clear outperformance, particularly in low-resource scenarios. This highlights the effectiveness of our data augmentation methods in enhancing ASR system accuracy and robustness, making them more suitable for handling diverse linguistic variations and real-world noise conditions.

Finally, The experimental results suggest that a well-designed augmentation strategy can bridge the gap between limited data availability and high model performance. This is crucial for developing ASR systems that are both accurate and versatile.

\section{Conclusions and Future Work} \label{sec:conclusions}

This paper aimed to improve an ASR model for under-represented languages with low data availability. We have demonstrated the effectiveness of fine-tuning the Wav2Vec2-xlsr-53 model on a fusion of augmented data for improving its performance in transcribing speech and capturing the meaning of the input audio. Moreover, our approach has outperformed the well-known and state-of-the-art Whisper model. A promising direction for future work may involve investigating the use of different types of augmentations, such as adding background noise or altering the speed or pitch of the speech. This could help us understand the effects of different augmentations on the performance of the Wav2Vec2-xlsr-53 model. Another direction is studying the impact of different amounts of augmented data on the performance of the Wav2Vec2-xlsr-53 model. This could help determine the optimal amount of augmented data needed to improve the model's performance. Lastly, exploring multilingual datasets where multiple languages are spoken in a single audio recording, such as in courts or inquiries, could provide valuable insights for developing more robust ASR systems.

\bibliographystyle{unsrtnat}
\bibliography{main}

\end{document}